\documentclass{ceurart} 

\usepackage{latexsym}
\usepackage{amssymb}
\usepackage{amsmath}
\usepackage{amsthm}
\usepackage{booktabs}
\usepackage{enumitem}
\usepackage{graphicx}
\usepackage{color}
\usepackage{array}
\usepackage{tabularx}
\usepackage{float}
\usepackage{multirow}
\usepackage{caption}
\usepackage[utf8]{inputenc}

\newcommand{\BibTeX}{B\kern-.05em{\sc i\kern-.025em b}\kern-.08em\TeX}
\newcolumntype{Y}{>{\raggedright\arraybackslash}X}

\begin{document}

\copyrightyear{2025}
\copyrightclause{Copyright for this paper by its authors.
  Use permitted under Creative Commons License Attribution 4.0
  International (CC BY 4.0).}

\conference{Edu4AI'25: 2nd Workshop on Education for Artificial Intelligence | co-located with ECAI 2025, Bologna, Italy}

\title{Investigating Bias: A Multilingual Pipeline for Generating, Solving, and Evaluating Math Problems with LLMs}

\author[1]{Mariam Mahran}[%
orcid=0009-0003-0568-0172,
email=mariam.mahran@htw-berlin.de,
url=https://mariamkhmahran.github.io/connect/,
]
\cormark[1]
\fnmark[1]

\author[1]{Katharina Simbeck}[%
orcid=0000-0001-6792-461X,
email=katharina.simbeck@htw-berlin.de,
url=https://iug.htw-berlin.de,
]
\cormark[1]
\fnmark[1]

\address[1]{HTW Berlin University of Applied Sciences, Treskowallee 8, 10318 Berlin, Germany}
\cortext[1]{Corresponding author}
\fntext[1]{These authors contributed equally.}


\begin{abstract}
Large Language Models (LLMs) are increasingly used for educational support, yet their response quality varies depending on the language of interaction. This paper presents an automated multilingual pipeline for generating, solving, and evaluating math problems aligned with the German K-10 curriculum. We generated 628 math exercises and translated them into English, German, and Arabic. Three commercial LLMs (GPT-4o-mini, Gemini 2.5 Flash, and Qwen-plus) were prompted to produce step-by-step solutions in each language. A held-out panel of LLM judges, including Claude 3.5 Haiku, evaluated solution quality using a comparative framework. Results show a consistent gap, with English solutions consistently rated highest, and Arabic often ranked lower. These findings highlight persistent linguistic bias and the need for more equitable multilingual AI systems in education.
\end{abstract}


\maketitle


\section{Introduction}
As Large Language Models (LLMs) become increasingly embedded in educational tools, they are reshaping how students access academic support \cite{Yan_2023, gan2023largelanguagemodelseducation, li2024adaptinglargelanguagemodels}. In subjects like mathematics, these models are often used to explain concepts, walk through problem-solving steps, and supplement classroom instruction \cite{Yan_2023}. Nonetheless, most commercial LLMs are trained on overwhelmingly English-centric data, raising concerns about how effectively they perform in other languages \cite{schut2025multilingualllmsthinkenglish, gupta2025multilingualperformancebiaseslarge}.

In multilingual educational settings, this imbalance can have serious implications. Students interacting with the same model in different languages may receive responses that vary not only in fluency, but in clarity, accuracy, and pedagogical value \cite{Yan_2023,ko2025understandsolvetranslatebridging}. This variation is especially problematic in math education, where precise explanations and terminology are essential. Prior studies have highlighted significant language-related disparities in LLM outputs, particularly in educational contexts \cite{ko2025understandsolvetranslatebridging,gupta2025multilingualperformancebiaseslarge}. These findings underscore the importance of systematic, scalable evaluation across languages.

In this paper, we present a scalable, automated pipeline for evaluating how language impacts the quality of LLM-generated math solutions. Covering generation, translation, solution, and evaluation, the framework supports consistent multilingual comparison across hundreds of exercises. Using three commercial models (GPT-4o-mini (OpenAI) \cite{openai2024gpt4omini}, Gemini-2.5-flash (Google) \cite{google2025gemini25flash}, and Qwen-plus (Alibaba Cloud) \cite{alibaba2025qwenplus}) we apply it to English, German, and Arabic to uncover persistent performance disparities in educational contexts.  
This work aims to drive future research toward reducing linguistic biases in AI education tools that serve learners regardless of language background.


\section{LLMs in education: Use Cases \& Biases}
LLMs are increasingly used in education to support both teachers and students. For educators, they assist with lesson planning, content creation, and assessment, helping reduce workload and improve efficiency \cite{Yan_2023,gan2023largelanguagemodelseducation}. For students, they offer real-time, adaptive support that fosters independent learning, clarifies difficult concepts, and guides problem-solving \cite{li2024adaptinglargelanguagemodels}.

Nonetheless, AI fairness in education is an increasing concern \cite{simbeck2024}. Growing evidence shows that LLMs introduce linguistic biases, particularly disadvantaging non-Western and low-resource language contexts \cite{gupta2025multilingualperformancebiaseslarge}. Models often default to Western entities (i.e. names, foods, locations...etc) even when operating in non-Western languages like Arabic \cite{naous2024havingbeerprayermeasuring}. Recent work also suggests that multilingual models often perform internal reasoning in English regardless of the prompt language \cite{schut2025multilingualllmsthinkenglish}. 

These tendencies raise concerns about the inclusivity and cultural relevance of AI-generated educational content. In mathematical reasoning tasks, LLMs have shown tendency to perform well in high-resource languages such as English and Chinese but struggle with languages like Korean due to difficulties processing non-English input \cite{ko2025understandsolvetranslatebridging}. Recent work by \cite{gupta2025multilingualperformancebiaseslarge} further demonstrate this gap by evaluating six LLMs on four educational tasks across six non-English languages. They find performance drops sharply in low-resource languages, correlating with training data representation. English prompts was also shown to outperform their translations across tasks. 

Such biases have real consequences. In grading and assessment contexts, models may favor Western argument structures, writing styles, and measurement systems placing non-Western students at a disadvantage. More broadly, the dominance of English-centric models reflects global educational inequality, as advanced AI tools remain most accessible to wealthier, English-speaking regions \cite{Yan_2023}.


\section{Dataset Creation}

\subsection{Dataset Generation \& Translation}

This study uses a dataset of 628 math exercises spanning five topic areas and grade levels B through H (equivalent to Grades 2-10), based on the official German K-10 mathematics curriculum \cite{RahmenlehrplanMathematik2015}.\footnote{Code and data are available at \url{https://github.com/iug-htw/solution_evaluator}.} Exercises were generated using ChatGPT-4o (via the ChatGPT interface) in a manual, iterative process guided by curriculum-aligned topic names and learning objectives. Rather than replicating specific examples, the curriculum served as a conceptual reference to ensure broad topic coverage and grade-level appropriateness. Each item was then manually reviewed for clarity, pedagogical relevance, and alignment with the intended mathematical concept. Unclear, repetitive, or misaligned exercises were revised or removed. The resulting dataset consists of brief, one-line prompts focused on single operations or concepts, minimizing the risk for any contextual or linguistic bias. Table~\ref{tab:exercise-examples} shows example exercises.

All exercises were initially written in English, then translated into German and Arabic using the GPT-4o-mini API. Given their concise and context-independent structure, the exercises were well-suited for machine translation. Nonetheless, 
all translations were manually audited to ensure semantic and linguistic appropriateness. Minor edits were applied as needed to maintain consistency across languages. 

\begin{table*}[t]
\centering
\caption{Examples of exercises across different topic areas and progress levels used in the study.}
\label{tab:exercise-examples}
\resizebox{\textwidth}{!}{
\begin{tabular}{p{3.5cm} p{3.9cm} p{1.9cm} p{6.1cm}}
\toprule
\textbf{Topic Area} & \textbf{Topic} & \textbf{Level} & \textbf{Exercise} \\
\midrule
Numbers and operations & Understanding and representing numbers & B (2nd grade) & Show 28 using tens and ones. \\
Sizes and measurement & units of measurement & C (4th grade) & Convert 4,289 mL to liters. \\
Geometry & measuring and calculating & F (8th grade) & A triangle has sides 7, 24, and 25 cm. Prove it is a right triangle. \\
Data frequencies and probabilities & Compare and evaluate data & E (7th grade) & Which measure (mean, median, mode) best represents this data: 3, 3, 3, 4, 100? \\
Functional relationships (patterns and structures) & Linear and other simple functions & H (10th) & Determine the x-intercepts and asymptotes of the logarithmic function f(x) = ln(x - 2) \\
\bottomrule
\end{tabular}}
\end{table*}

\subsection{Solution Generation}
Three LLMs were used in this study: \textit{GPT-4o-mini}, \textit{Gemini 2.5 Flash}, and \textit{Qwen-plus}. Each was prompted to generate step-by-step solutions in English, German, and Arabic. The prompts, written in the target language, asked the model to explain how to solve the given problem rather than to provide a final answer. This setup was intended to mirror how K-10 learners might seek guided support or clarification. The resulting solutions were saved in separate CSV files for each language-model combination.


\section{Evaluation}

To evaluate solution quality, we adopted a hybrid evaluation framework in which \textbf{LLMs acted as judges}. LLM-based evaluation is gaining traction as a scalable alternative to human or code-based assessments \cite{zheng2023judgingllmasajudgemtbenchchatbot}. Findings suggest that LLMs can generate assessments comparable to human judgments in domains such as open-ended story generation and adversarial attack evaluations \cite{chiang-lee-2023-large}. In educational contexts, LLM judgments have shown strong correlation with faculty evaluations, particularly when using pairwise comparisons and structured rubrics \cite{ishida2024largelanguagemodelspartners}. Other studies have shown that, with appropriate fine-tuning and prompt design, LLMs can effectively automate course evaluations across diverse university settings  \cite{yuan2024explorationhighereducationcourse}.

For each math exercise, three step-by-step solutions, one in English, German, and Arabic, were presented as alternative responses to the same prompt. These were then assessed and compared by a panel of LLM judges based on their clarity, accuracy, and pedagogical effectiveness.

To ensure impartiality and prevent indirect bias from models evaluating their own outputs, we implemented a \textbf{held-out judging strategy}. In each round, the model under evaluation was excluded from the judging panel and replaced with a neutral model, \textit{Claude 3.5 Haiku} \cite{anthropic2025claude35}. The remaining three judges were selected from \textit{GPT-4o-mini}, \textit{Gemini-2.5-flash}, \textit{Qwen-plus}, and \textit{Claude}. 

Prior to evaluation, we identified technical terms relevant to each exercise, as the accurate use and explanation of such terms are crucial for clarity and learner comprehension. To do so, we employed a script that extracted these terms from each exercise using the \textit{GPT-4o-mini} model.

During evaluation, solutions were presented in \textbf{randomized order} to each judge to mitigate position bias. Position bias occurs when the order of presentation influences rankings, a phenomenon observed both in LLMs and human decision-making \cite{tan2024judgebenchbenchmarkevaluatingllmbased, zheng2023judgingllmasajudgemtbenchchatbot}.

We followed a \textbf{comparative assessment} methodology, which has been shown to outperform direct scoring in terms of alignment with human preferences \cite{liu2025aligninghumanjudgementrole, liusie2024llmcomparativeassessmentzeroshot}.  Judges were asked to rank the three solutions from 1 (best) to 3 (worst) and provide a short justification to their judgment. Rankings were based on multiple educational criteria: conceptual understanding, clarity of explanation, step-by-step reasoning, correct terminology, accuracy, recognition of common mistakes, pedagogical suitability, generalizability, and grade-level appropriateness.

To identify the best-performing language for each exercise, we applied a \textbf{majority voting} scheme. If two or more judges ranked the same solution highest, the language of that solution was recorded as the top performer.  If the judges couldn't agree on a single best-performing solution, the result was labeled a tie (TIE), ensuring that ambiguous cases were recorded without forcing a winner.

\begin{table*}[t]
\centering
\caption{Comparison of solutions ranking across the three models.}
\label{tab:combined-results}
\renewcommand{\arraystretch}{1.2}
\begin{tabularx}{\textwidth}{l *{9}{>{\centering\arraybackslash}X}}
\toprule
 & \multicolumn{3}{c}{\textbf{GPT-4o-mini}} & \multicolumn{3}{c}{\textbf{Gemini-2.5-flash}} & \multicolumn{3}{c}{\textbf{Qwen-plus}} \\
\cmidrule(lr){2-4} \cmidrule(lr){5-7} \cmidrule(lr){8-10}
\textbf{Language} & Best & Mid & Worst & Best & Mid & Worst & Best & Mid & Worst \\
\midrule
English & 493 & 80  & 19  & 233 & 233 & 138 & 254 & 206 & 152  \\
German  & 76  & 325 & 167 & 184 & 180 & 246 & 194 & 195 & 228  \\
Arabic  & 38  & 131 & 403 & 206 & 188 & 223 & 174 & 199 & 239 \\
TIE     & 21  & 92  & 39  & 5   & 27  & 21  & 6   & 28  & 9   \\
\bottomrule
\end{tabularx}
\end{table*}


\section{Results}

Table \ref{tab:combined-results} summarizes the performance of solutions per language, categorizing them for each LLM based on how frequently they were ranked as the best, mid-tier, or worst. Across all three models, English solutions were most frequently favored. GPT-4o-mini showed the strongest language preference, heavily favoring English and consistently rating Arabic lowest. Qwen-plus followed a similar trend, though with less extreme separation. Gemini-2.5-flash produced a more balanced distribution overall; while English still led in top rankings, the difference between German and Arabic was narrower, and German received the most “Worst” labels in that case.

In addition to collecting quantitative rankings, we asked the LLM judges to provide brief justifications for each evaluation (see examples in Table \ref{tab:explanations}). Analysis of over 4,000 justifications revealed consistent emphasis on clarity, structured reasoning, and appropriate use of mathematical terminology. English solutions were most frequently praised as “comprehensive,” “well-structured,” or “clear,” reflecting strong positive sentiment overall. German responses received more mixed feedback, often described as adequate but occasionally critiqued for being “less detailed” or assuming prior knowledge. Arabic solutions had the highest concentration of negative sentiment, with justifications noting a lack of clarity or depth in explanation. Notably, comparative framing (e.g., “slightly better,” “strongest”) was common, confirming that judges made relative, not absolute, assessments.


\section{Discussion}

The results indicate a clear disparity in performance across languages. English solutions were rated most frequently the best, while Arabic was generally rated the worst, although not uniformly across models. German solutions typically occupied the middle range. GPT-4o-mini showed the most extreme language separation, assigning overwhelmingly high scores to English and low scores to Arabic. Gemini-2.5-flash diverged slightly, assigning the highest number of “Worst” rankings to German rather than Arabic. However, the margin between German and Arabic was relatively small (246 vs. 223), and Gemini’s rankings were more evenly distributed across languages overall. 

It is worth noting that, while we did not repeat the full experiment under fixed conditions, the full pipeline was executed multiple times during development with variations in datasets, prompts, and code refinements. Despite fluctuations in exact scores (as expected with stochastic outputs), the relative pattern remained consistent: English was favored, Arabic was disadvantaged, and German remained in between. This recurring trend across settings suggests that the observed disparities are robust rather than incidental.

These gaps were not solely driven by solution correctness. Even when all three solutions were mathematically accurate, models showed clear preferences for responses that included explicit reasoning, pedagogical structure, and generalizability. Analysis of the provided justifications show that the LLMs consistently reward explanations that go beyond getting the answer right to showing \textit{why} the method works.  Another key insight is that each language showed strengths in different areas. Arabic was often rewarded for its warmth and accessibility, especially at lower grade levels. German performed well when structure and terminology were key but sometimes assumed prior knowledge. These patterns suggest that language-specific conventions shape both how explanations are presented and how models evaluate their educational value.

Nonetheless, variations in linguistic complexity likely contributed to the observed performance patterns. German has a complex sentence structure with long compound words, flexible word order, and grammatical rules. These characteristics can make mathematical expression and phrasing more difficult for LLMs to manage, which may explain its intermediate performance. Arabic with its rich morphology and right-to-left script posed even greater challenges. This morphological richness can compress tokens, producing shorter outputs that feel abrupt next to more elaborated English or German responses. Moreover, LLMs may also lack exposure to Arabic instructional texts with formal, school-style structure. As a result, LLMs may misinterpret Arabic's concise or direct solutions as less educationally complete, even when correct. These surface-level differences can shape how LLMs perceive clarity and instructional value, even when conceptual accuracy is maintained.

The decision to fully automate the pipeline was driven by the need for scalable, reproducible analysis. Manual evaluation methods, while valuable, limits consistency and scope, making it difficult to compare results at scale. Our automated approach ensures uniform treatment across hundreds of exercises and enables easy replication or extension to new models, languages, or curricula. It also reduces human bias in assessment and supports more systematic multilingual benchmarking.

Finally, the findings carry clear pedagogical implications. Teachers cannot assume equal support across languages, as LLMs may provide stronger guidance in English than other languages. LLMs should therefore be treated as supplementary tools, ideally with teacher oversight or cross-lingual checks. At the same time, the pipeline itself serves as an analytical resource, helping educators pinpoint curriculum areas where linguistic gaps pose the greatest risk, such as topics requiring precise terminology or multi-step reasoning. This allows teachers to adapt instruction accordingly to promote more equitable learning in multilingual classrooms.

\begin{table}[t]
\centering
\caption{Examples of LLM ranking justifications for each language}
\label{tab:explanations}
\renewcommand{\arraystretch}{1.3}

\resizebox{\textwidth}{!}{
\begin{tabular}{p{8.15cm} p{8.15cm}}
\toprule
\textbf{Positive Feedback} & \textbf{Negative Feedback} \\
\midrule
\textnormal{[en]}: "...\textbf{well-structured} and clearly explains the concept using \textbf{appropriate terminology} and visual analogies."  
& \textnormal{[en]}: "...while correct, \textbf{lacks some of the engaging elements}...may not connect with younger students" \\

\textnormal{[de]}: "...excels in \textbf{clarity}, \textbf{structure}, and appropriate \textbf{use of technical terms}...avoids assumptions..." 
& \textnormal{[de]}: "... although correct, \textbf{assumes prior knowledge} of rearranging the density formula..." \\

\textnormal{[ar]}: "...provides a \textbf{student-friendly approach} to solving the problem...making it easy for a 4th grader to follow."  
& \textnormal{[ar]}: "... presents the information in a \textbf{slightly less structured} manner..." \\

\bottomrule
\end{tabular}}
\end{table}


\section{Conclusion}
This study presents a scalable, fully automated pipeline for evaluating LLM-generated math solutions across languages. Applying it to English, German, and Arabic revealed consistent disparities in explanation quality, with English generally favored and Arabic more frequently disadvantaged. These differences were not solely about correctness but reflected variation in reasoning depth and pedagogical clarity. 

As LLMs become increasingly integrated into classrooms and tutoring platforms, such disparities risk reinforcing existing educational inequalities. Our findings underscore the importance of more linguistically inclusive development and evaluation practices. To ensure equitable access, future work should prioritize improving performance in underrepresented languages through targeted fine-tuning, diverse training data, and culturally informed prompt design. Expanding this evaluation framework to additional languages and domains can support the development of more accessible and fair educational AI systems worldwide.

\begin{acknowledgments}
This paper portrays the work carried out in the context of the KIWI project (16DHBKI071) that is generously funded by the Federal Ministry of Research, Technology and Space (BMFTR).
\end{acknowledgments}

\section*{Declaration on Generative AI}
During the preparation of this work, the authors used OpenAI’s GPT-4.1 to check grammar and spelling and to enhance the writing style. After using these tools/services, the authors reviewed and edited the content as needed and take full responsibility for the publication’s content.


\bibliography{mybibfile}

\end{document}